# An Aggregation of Aggregation Methods in Computational Pathology

Mohsin Bilal[1], Robert Jewsbury [1], Ruoyu Wang[1], Hammam M. AlGhamdi[1], Amina Asif[1], Mark Eastwood[1], Nasir Rajpoot[1,2,3]✉

[1] Tissue Image Analytics Centre, Department of Computer Science, University of Warwick, UK
[2] The Alan Turing Institute, UK
[3] Department of Pathology, University Hospitals Coventry and Warwickshire, UK

✉ Corresponding Author. Email: N.M.Rajpoot@warwick.ac.uk, Tel: +44 (24) 7657 3795

**Abstract**

Image analysis and machine learning algorithms operating on multi-gigapixel whole-slide images (WSIs) often process a large number of tiles (sub-images) and require *aggregating* predictions from the tiles in order to predict WSI-level labels. In this paper, we present a review of existing literature on various types of aggregation methods with a view to help guide future research in the area of computational pathology (CPath). We propose a general CPath workflow with three pathways that consider multiple levels and types of data and the nature of computation to analyse WSIs for predictive modelling. We categorize aggregation methods according to the context and representation of the data, features of computational modules and CPath use cases. We compare and contrast different methods based on the principle of multiple instance learning, perhaps the most commonly used aggregation method, covering a wide range of CPath literature. To provide a fair comparison, we consider a specific WSI-level prediction task and compare various aggregation methods for that task. Finally, we conclude with a list of objectives and desirable attributes of aggregation methods in general, pros and cons of the various approaches, some recommendations and possible future directions.

## 1. Introduction

The emerging area of computational pathology (CPath) involves a broad range of computational methods to analyse digitized images of tissue slides for a wide variety of downstream applications such as clinical decision-making and biomarker analysis (Abels et al., 2019). A high-resolution scan of a routine histology slide of a tissue specimen generates a whole slide image (WSI), often containing several billions of pixels. A WSI is a multi-gigapixel image containing large amount of information-rich pixel data at various levels of details, for instance a large number of various types of cells and glands, tissue phenotypes, and regions of interest to analyse for WSI-level predictions. However, a WSI can often not be processed entirely in graphic processing units (GPUs) for training or inference, presenting a computational challenge in itself. As a remedy, it is common to divide a WSI into multiple image tiles (or patches), perform the analysis on individual patches (or small groups of patches) and *aggregate* the results of inference from various levels into a WSI-level inference. While localization and recognition of objects like cells benefits from low-level details of the WSI data, prediction at the level of image tiles themselves provides slightly better context at the cost of missing low-level details of the cellular objects, a manifestation of the classical position-class uncertainty trade-off (Wilson & Knutsson, 1988). In this survey, we focus on computational approaches to WSI analysis for various different WSI-level tasks including prediction of diagnostic and molecular labels and survival analysis.

There are multiple levels of aggregation in CPath. For example, aggregation of object (say, nuclei) level predictions into an image patch level prediction and aggregation of patch-level predictions into larger tile level prediction, as illustrated in Figure 1a. In this paper, we focus on methods that aggregate predictions gathered from objects, image patches or tiles into WSI-level predictions. These methods follow the pipeline shown in Figure 1b. As part of our comprehensive review, we aim to cover all aspects of CPath literature, including data in a WSI, computational approaches, use cases, evaluation, comparison, and recommendations for possible future directions.

Generally speaking, a CPath model's WSI-level inference uses an aggregated score to label a WSI with a correct diagnostic, prognostic or molecular category, making *aggregation* an essential module in several CPath analytical pipelines. We would like to note that our definition of "aggregation" in this paper is not restricted to aggregated scores only. It also refers to aggregating features obtained from various levels, objects and parts of a WSI for predictive modelling. For instance, there can be more than one WSIs for a patient to model CPath solutions, which require aggregating scores from multiple WSIs per case (Chang et al., 2021).

## 1. Computational pathology workflow

Figure 1 illustrates whole slide image analysis workflow to approach the predictive modelling solution by processing the WSIs in three different ways.

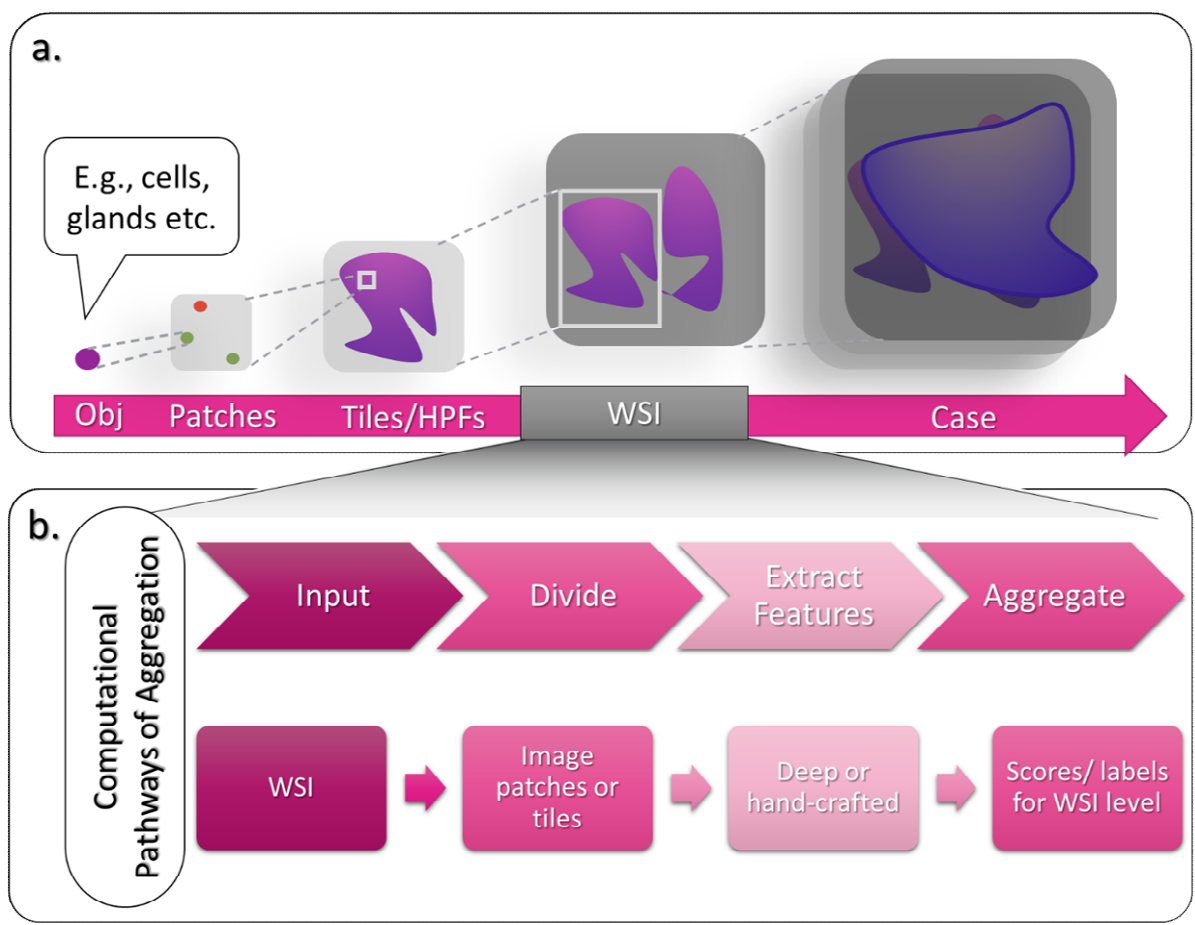

*Figure 1. A general overview of computational pathology data and workflows, a. A WSI contains many cells and glands as objects (Obj.), image patches, tiles or high power fields (HPF). A case may have multiple WSIs. b. Whole slide image analysis workflow here, considers a WSI as input, divided into several image tiles (patches or HPF) for machine learning to compute an aggregated output at WSI-level. c. Image tiles from a WSI can be processed by machine learning model to infer features; i.e., scores or labels or deep features, to be aggregated by different aggregation methods.*

A CPath model obtains image tiles by dividing WSIs and WSI-level scores by aggregating the results of inference on the tiles. The predictive modelling in a CPath pipeline may follow one of the three approaches: bottom-up inference, top-down inference or tissue phenotypic representation based inference. In the bottom-up approach, image patches or tiles are used to detect, segment and classify various tissue objects like cells and glands as the primary units of information that are then subject to aggregation to represent a WSI, see for instance (Diao et al., 2021; Ho et al., 2022; W. Lu et al., 2020; Park et al., 2022). The WSI representations for predictive modelling offer options from classical machine learning to graph learning (GL) where graph convolutional neural networks (GCNNs) learn and aggregate all information into a single score for clinical decision-making.

The top-down approach begins with analysing WSIs with tile-level or region-level predictions. In its simpler form, it does not require any annotations for specific objects and regions in a WSI, instead a WSI-level label may be used to weakly label the small image tiles. Tile-level scores or deep features can then be used to model predictive analysis, for instance in a multiple instance learning (MIL) setting. Results of inference on tiles are then aggregated into a WSI-level score (Bilal et al., 2021; Coudray et al., 2018; Kather, Pearson, et al., 2019).

There is a third approach in CPath workflows, in which we first learn to distinguish different tiles as tissue phenotypes or regions of interest (labelled) within a WSI to incorporate *apriori* domain knowledge in predictive modelling (Park et al., 2022; Su et al., 2022; K. S. Wang et al., 2021; Yamashita et al., 2021), leading to a tissue phenotypic representation of the WSI.

Figure 2 and Table 1 present a summary of the aggregation methods found in the CPath literature considering three related aspects of the scientific literature: use cases, the type of input data and aggregation methods.

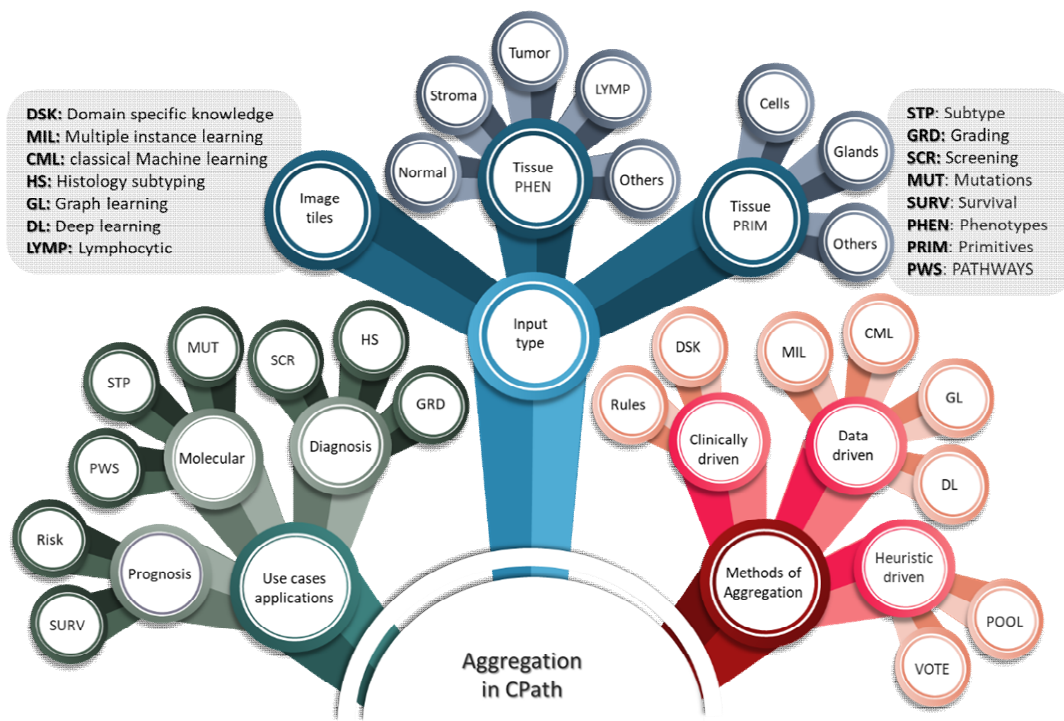

*Figure 2. Aggregating the aggregation literature: use cases, input type and methods of aggregation.*

In this paper, we focus on WSI image analysis workflows for the diagnostic, molecular and prognostic survival predictions at the WSI level as three main histopathology use cases. We have also added an aggregation method which used multiple WSIs and generated an aggregated output at the case level. Table 1 provides more details on the methods for these use cases with additional information of cancer type, clinical problem, aggregation method and workflows, and datasets used whether public or private. Diagnostic tasks include primary cancer screening, cancer subtype classification, cancer grade prediction and metastases detection. In the molecular prediction tasks, we include workflows for gene expression prediction, molecular pathways/subtype prediction, mutation prediction and treatment response prediction (Xie et al., 2022). In the prognostic use case, we include survival and risk prediction workflows. In the next section, we describe methods of aggregation in detail.

# 1. Methods of Aggregation

Most CPath approaches process WSI tiles and aggregate the tile-level predicted labels, scores or probabilities to predict the slide-level label. We refer to both the small image patches and large image tiles as *tiles* from hereon. The problem can be formulated as follows: given a WSI $X$ composed of a set or a *bag* of tiles $X = \{x_1, x_2, \ldots, x_n\}$ and their corresponding predictions $y = \{y_1, y_2, \ldots, y_n\}$, the output prediction $y_{wsi}$ is obtained as follows,

$$y_{wsi} = g_\varphi(\{y_i = f_\theta(x_i), \forall i\}) \qquad (1)$$

In equation (1), $f$ represents a function (usually a neural network) that converts a patch $x_n$ into an instance level probability, score or prediction and is parameterised by $\theta$. The function $g$ is the aggregation function that takes the predictions of $f_\theta(x_i), \forall i$ and combines them into a slide-level prediction. It can have learnable parameters of its own, denoted by $\varphi$, or it can be non-parameterised.

Broadly speaking, three types of approaches for defining the function $g$ can be found in the literature, as described in the remainder of this section.

## 1.1. Heuristic/Statistical Aggregation

Heuristic approaches also known as global pooling approaches require no machine learning and are not data dependent. Each instance is processed independently and then the scores are aggregated using a fixed formula. All the pooling approaches condense a set of tile-level scores into a slide-level score in some way. There are many different formulations for $g(\cdot)$ but the most commonly used ones are:

- Mean: $g(y) = \frac{1}{n}\sum_{i=1}^{n} y_i$ where $n$ is the number of tiles in the WSI.
- Max: $g(y) = \max(y_i); i = \{1, 2, \ldots, n\}$

Others include Majority voting (mode), Generalised mean (Y. Xu et al., 2014), Noisy-OR (Kraus et al., 2016) and Noisy-AND (Skrede et al., 2020).

## 1.1. Data-driven Aggregation

In data-driven approaches, the function $g$ has a set of learnable parameters $\varphi$. The most commonly used approach is a form of attention aggregation proposed by Ilse *et al.* (Ilse et al., 2018), which takes a set of feature vectors $z$ extracted from the input bag of instances $X$ using a neural network $f$ parameterised by $\theta$

$$z = f_\theta(X) = \{z_1, z_2, \ldots, z_n\} \qquad (2)$$

The bag/slide label is obtained via the following formulation,

$$y_{wsi} = g(z) = \sum_{i=1}^{n} a_i z_i \qquad (3)$$

*Table 1 Summary of aggregation methods in CPath*

| Authors | Cancer type(s) | Clinical Problem(s) | Aggregation Method |
|---|---|---|---|
| *Non-data driven methods* | | | |
| **Schmauch et al (2020)** (Schmauch et al., 2020) | 28 cancer types | Gene expression | Weighted average pooling |
| **Gildenblat et al (2021)** (Gildenblat et al., 2021) | Breast cancer | Metastases detection | Certainty Pooling |
| **Skrede et al (2020)** (Skrede et al., 2020) | Colorectal cancer | Cancer specific survival | Noisy And Pooling |
| **Bilal et al (2021)** (Bilal et al., 2021) | Colorectal cancer | Molecular pathways / Mutation prediction | Iterative draw and rank sampling (IDARS), average probability aggregation |
| **Yamashita et al (2021)** (Yamashita et al., 2021) | Colorectal cancer | Microsatellite instability prediction | Tissue phenotype classification and average probability aggregation |
| **Bilal et al (2022)** (Bilal et al., 2022) | Colorectal cancer | Cancer screening | IDARS, Average (of tiles with probability greater than median) probability aggregation |
| **Kather et al (2019)** (Kather, Pearson, et al., 2019) | Colorectal / Gastric cancer | Molecular subtypes / Mutation prediction | Naive MIL, Proportion of positive predicted tiles |
| **Su et al (2022)** (Su et al., 2022) | Gastric cancer | Microsatellite instability recognition | Majority voting |
| **Kanavati et al (2020)** (Kanavati et al., 2020) | Lung cancer | Sub-type classification / Metastases detection | Max pooling |
| *Parameterised methods* | | | |
| **Diao et al (2021)** (Diao et al., 2021) | 5 cancer types | Molecular phenotype prediction | Classical machine learning based aggregation |
| **Chen et al (2022)** (R. J. Chen et al., 2022) | 8 cancer types | Sub-type classification / Survival prediction | Hierarchical transformer aggregation |
| **Lipkova et al (2022)** (Lipkova et al., 2022) | Allograft rejection | Rejection conditions / Grade prediction | Multitask attention MIL |
| **Hashimoto et al (2020)** (Hashimoto et al., 2020) | Blood cancer | Sub-type classification | Domain adversarial attention MIL |
| **Lu et al (2021)** (M. Lu et al., 2021) | Brain cancer | Sub-type classification | Contrastive and sparse-attention based MIL |
| **Sharma et al (2021)** (Sharma et al., 2021) | Breast / Gastric cancer | Metastasis detection / Celiac prediction | Clustering and Attention MIL |
| **Lu et al (2021)** (M. Y. Lu et al., 2021) | Breast / Kidney / Lung cancer | Sub-type classification / Metastases detection | Attention MIL and Instance-level clustering |
| **Li et al (2021)** (Li et al., 2021) | Breast / Lung cancer | Sub-type classification / Metastases detection | Dual instance and bag aggregation |
| **Campanella et al (2019)** (Campanella et al., 2019) | Breast / Prostate / Skin cancer | Metastases detection / Grade prediction | Top-k learning, RNN aggregation |
| **Lu et al (2022)** (W. Lu et al., 2022) | Breast cancer | Mutation prediction (HER2) | GNN aggregation |
| **Naik et al (2020)** (Naik et al., 2020) | Breast cancer | Mutation prediction | Attention MIL |
| **Tellez et al (2019)** (Tellez et al., 2019) | Breast cancer | Metastases detection / Tumour proliferation speed | Neural Image Compression |
| **Schirris et al (2022)** (Schirris et al., 2022) | Breast cancer / Colorectal cancer | Mutation prediction | SSL encoder and VarMIL |
| **Shao et al (2021)** (Shao et al., 2021) | Breast / Kidney / Lung cancer | Sub-type classification / Metastases detection | Transformer aggregation |
| **Park et al (2022)** (Park et al., 2022) | Colorectal / Gastric cancer | Microsatellite instability prediction | Mean aggregation & light gradient boosting machine |
| **Saillard et al (2021)** (Saillard et al., 2021) | Colorectal / Gastric cancer | Microsatellite instability prediction | SSL encoder, top and bottom scores and Chowder |
| **Reisenbuchler et al (2022)** (Reisenbüchler et al., 2022) | Colorectal / Stomach cancer | Mutation prediction | Local attention graph transformer |
| **Ho et al (2022)** (Ho et al., 2022) | Colorectal cancer | Cancer detection | Gland segmentation and aggregation by gradient-boosted decision tree |
| **Tomita et al (2019)** (Tomita et al., 2019) | Esophageal cancer | Sub-type classification | Attention MIL |
| **Xie et al (2022)** (Xie et al., 2022) | Lung cancer | ICI treatment response prediction | End-to-end part-learning GNN |
| **Chang et al (2021)** (Chang et al., 2021) | Lung cancer | Survival analysis | Hybrid aggregation network |
| **Pinckaers et al (2020)** (Pinckaers et al., 2020) | Prostate cancer | Grade prediction | Streaming CNN |
| **Zhang et al (2022)** (Zhang et al., 2022) | Rectal cancer | Chemoradiotherapy efficacy prediction | Multi-scale CNN bilinear attention MIL |

where

$$a_i = \frac{\exp\{\underline{w}^T \tanh(\underline{V} z_i^T)\}}{\sum_{j=1}^{n} \exp\{\underline{w}^T \tanh(\underline{V} z_j^T)\}} \tag{4}$$

and $\underline{w} \in \mathbb{R}^{L \times 1}$ and $\underline{V} \in \mathbb{R}^{L \times N}$ are learnable parameters. It is worth noting that this formulation assumes all instances are independent.

There are many other formulations that have been proposed including extensions of this attention MIL (Lipkova et al., 2022; M. Y. Lu et al., 2021), variations that use positional encodings in the attention structure (Shao et al., 2021), ones that use a Recurrent Neural Network (RNN) (Campanella et al., 2019) or a Transformer (R. J. Chen et al., 2022) as the aggregation function instead of the attention mechanism. Gildenblat *et al.* (Gildenblat et al., 2021) proposed a data-driven mechanism of certainty pooling in the MIL framework showing better performance than attention mechanism and heuristic aggregation like mean and max pooling.

Other approaches for the data-driven aggregation includes graph learning (W. Lu et al., 2020) and the application of classification machine learning like random forest, logistic regression, support vector and gradient boosting machines with the handcrafted features of WSI obtained from the tiles, cells, and glands (Diao et al., 2021; Ho et al., 2022). Handcrafted features take the tile-level prediction scores or probabilities in the same way as the prior two methods but instead of aggregating these to the prediction for the WSI level label directly they extract features from the tile scores such as a histogram of tile prediction scores. Other examples include morphological or statistical features extracted from the prediction maps.

### 1.2. Clinically-Driven Aggregation

The final class of aggregation methods use clinical formulae developed by pathologists and currently used in clinical tasks. For example, in breast cancer to categorise the amount of HER2 receptor protein on the surface of cells in the sample, the Royal College of Pathologists (RCPath) guidelines suggest specific criteria as defined in RCPath report, p.99 (Ellis et al., 2016). Other examples of aggregation using domain specific knowledge include mismatch repair assessment to determine microsatellite instability (MSI) status of the sample using the expression of immunohistochemistry (IHC) with 4 antibodies (MMR-IHC: MLH1, MSH2, MSH6 and PMS2) (Awan et al., 2022) and assessment of PD-L1 protein expression using tumour proportion score (Pagni et al., 2020). The latter approach detects and quantifies tumour cells in a WSI in terms of the amount of IHC stain that has been absorbed and then compute the percentage of tumour cells that are strongly stained and compare it with these criteria.

## 2. Multiple instance learning

Multiple instance learning or MIL is a paradigm of supervised machine learning that deals with incomplete and ambiguous information of labels in training data. The learner receives a set of labelled *bags* where each bag has multiple unlabelled instances. Dietterich *et al*. (Dietterich et al., 1997) first coined the term. In its basic form, the MIL problem restricts to a binary classification of bags (Babenko, 2008), but other forms of multiple instance regression (Ray, 2001) and multi-instance multi-label learning (Z.-H. Zhou & Zhang, 2007) can also be found in the machine learning literature. The basic MIL assumption is that every positively labelled bag contains at least one positive "witness" or "key" instance (Babenko, 2008). It implies that all instances in a negatively labelled bag are negative instances. Babenko in (Babenko, 2008) presents a thorough review of different MIL methods from classical machine learning for further reading. We have choices of modelling the MIL problem as an instance classifier, a bag classifier (Babenko, 2008) or a combination of instance and bag classifiers.

Predictive modelling in CPath is analogous to the MIL problem if it considers the WSI as a bag with a single label for the purposes of predictive modelling, which is often the case in CPath workflows. Having multiple classes and several labels for each WSI (or bag) is also possible in CPath problems, where the basic assumption of the MIL problem may violate. For some CPath problems, bags may also have instances irrelevant to the prediction task. In other words, both the positive and negative bags may have noisy samples, which are not related to the given label of the WSI. An alternate term of weakly-supervised learning might be more appropriate in such scenarios.

### 2.1. Multiple instance learning in CPath

In this section, we review recent well-known MIL methods in CPath. Figure 3 illustrates six recently published MIL methods, which consider a WSI as a bags of patches. We group these methods according to their representation, learning and aggregation modules.

A naive MIL approach fine-tunes all or a few of the last layers of a convolutional neural network (CNN), InceptionV3 in (Coudray et al., 2018) and ResNet18 in (Kather, Pearson, et al., 2019), pre-trained on ImageNet. Each patch gets the same label as the bag (or WSI) during the model's fine-tuning, which means all patches in the positive bag get the positive label. It adds a potential noise in the model training by forcing the network to learn irrelevant or negative instances as positive. The model training follows an instance classifier approach. The model testing applies an average or a majority voting scheme to aggregate probabilities of image patches into scores of a bag or WSI. The majority vote measures a ratio of positively predicted instances over all instances in the bag. Several publications have used naive MIL approach to predict diagnostic and molecular labels of WSIs.

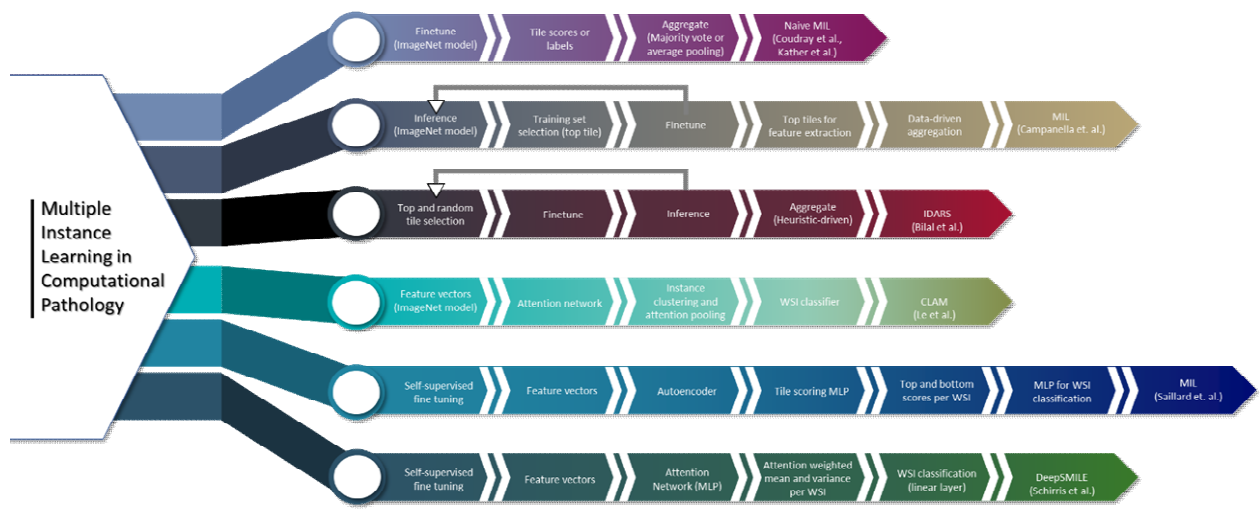

*Figure 3. Popular multiple instance learning pipelines in computational pathology*

Campanella *et al*. (Campanella et al., 2019) proposed an advanced MIL approach for clinical-grade diagnosis of prostate cancer, basal cell carcinoma and breast cancer metastasis. During training, they fine-tune a pretrained CNN (ResNet34) before the CNN is used to predict a score for each tile and then only the top tiles of positive class in each WSI are used for *training*. In the first stage, this pair of *inference* followed by *training* recurs for maximum number of iteration (e.g. 100) to obtain a trained CNN. In the second stage, the trained CNN predicts a few top (e.g. 20) positive tiles from each slide for training an aggregation network. As an aggregation network, they train recurrent neural network on last layer features of selected top tiles and compare it with a random forest trained on the hand-crafted feature of top twenty tiles. This was shown to achieve clinical-grade performance for binary diagnostic tasks when a large number of WSIs were used for training the model.

Clustering-constrained attention multiple instance learning (CLAM) (M. Y. Lu et al., 2021) is another MIL based aggregation method that represents each WSI as a fixed size bag and each input image patch using fixed features of ResNet50 pre-trained on ImageNet. It performs data-driven aggregation, which involves multi-class attention-based learning to identify associated diagnostic subregions to accurately classify WSI and instance-level clustering over the identified representative regions to constrain and refine the feature space. Several recent studies have employed this method for WSI label predictions.

Bilal *et al*. proposed iterative draw and rank sampling (IDaRS) (Bilal et al., 2021) for fine-tuning a CNN (ResNet34) on two smaller subset (random (*r*) and top (*k*)) of tiles from each WSI. After initiating the training with random tiles (50 or 11%), they obtain top *k (5 or 1%)* positive tiles in each subsequent iteration to combine with random (*r*) tiles from each WSI for training. For aggregation, they experiment with several pooling methods like average of positive probability of all tiles or selective tiles (top few (5 or 10), top half (50%)), and weighted average. They report better AUROCs and average precision of PR-curves with average and top half aggregation for molecular (Bilal et al., 2021) and diagnostic (Bilal et al., 2022) labels predictions.

Recently, self-supervised learning based (Saillard et al., 2021), (Schirris et al., 2022) have been proposed for various histopathology tasks. Saillard *et al*. (Saillard et al., 2021) use self-supervised learning to fine-tune a pretrained CNN (ResNet50). For each tile, they use an autoencoder to reduce the last layer features of trained ResNet50 to a 256-dimensional vector. They predict microsatellite instability with three different MIL frameworks for aggregation. A better performing aggregation network consists of two multilayer perceptron (MLP) networks. The first MLP processes features to score each tile for the selection of $R$ ($R$=10, 25, or 100) top and bottom scores per WSI for the training of second MLP. The second trained MLP processes concatenated top and bottom scores to infer a final aggregated score. In first stage of the DeepSMILE (Schirris et al., 2022), authors use self-supervised learning for fine-tuning a pre-trained CNN (ResNet18 and ShuffleNetV2). In second stage, DeepSMILE proposes MLP-based aggregation network. The aggregation network uses the last layer features of the trained CNN and learns attention followed by a classification module to get an aggregated output for the prediction of MSI and homologous recombination deficiency (HRD).

## 3. Context in aggregation

In visual processing systems, data-driven modelling requires visual context without losing an appropriate level of finer details. In CPath, processing bags of image tiles without spatial information compromises the wider visual context. Capturing visual context with constrained computational resources results in losing the finer details and raises a trade-off between visual details and context. If we retain a fuller context at the lowest level of resolution, we will lose essential details of the WSI that impedes predictive modelling. In its trivial form, MIL-based approaches described in Section 4.1 considered patching at an appropriate resolution but with a limited context and without retaining the spatial information of image patches. The aggregation attempts to recover the lost context from disjoint patches in the bag. Yet, this may be a suboptimal solution until we can process an entire WSI as a single sample or retain deep features besides spatial location and awareness of their role in the final prediction. Another approach often found in the literature analyses patches at multiple resolutions, which can be beneficial to capture the heterogeneity of data from multiple regions (Zhang et al., 2022). Next, we describe the global context and graph-based aggregation approaches attempting WSI image analysis with a global context and spatial interaction.

## 3.1. Global context

Global context aggregation approaches attempt to address the limitations of local context approaches where the narrow field of view considered in the tile aggregation approaches limits the incorporation of global features. The typical tile size in computational pathology problems is 256x256 or 512x512 pixels. Particularly at higher magnifications, e.g. 20 or 40x, this results in a bag of instances with no consideration for the spatial relationship between the different instances. Efforts to address this problem outside of graph structures are few but varied.

### 3.1.1. Transformers

The Transformer architecture (Vaswani et al., 2017) has quickly become the state-of-the-art for many language tasks and recently with the introduction of the Vision Transformer (Dosovitskiy et al., 2021) has been applied extensively for computer vision tasks as well. It uses a multi-head self-attention mechanism which unlike RNNs does not consider the order or relative position of tokens in an input sequence. To address this, the transformer uses a positional encoding with each input token adding in the relative position of each component in the sequence. This allows the Transformer to have greater awareness of longer-range dependencies in the input data compared to other models like a CNN's kernels that have a fixed window size. As such it lends itself to address the lack of context problem associated with the local aggregation methods.

To aggregate image tile instances for WSI classification, several methods using Transformers have been used such as SMILE (M. Lu et al., 2021), TransMIL (Shao et al., 2021) and GTP (Zheng et al., 2021). They operate on a similar paradigm to the local context approaches but attempt to encode in the feature vectors an aspect of the given feature vector's spatial positioning with respect to other feature vectors. The construction of a bag of instances is usually handled in the same way as the local context methods, ie a tissue mask is used to extract only tissue tiles which are then fed through a pre-trained or a fine-tuned encoder, usually a ResNet50. It is after this step that the Transformer approaches differ. SMILE uses a SAM (Sparse-Attention) module to extract the top-N feature embeddings from the bag of instances which is then fed forward into a transformer module. TransMIL uses a Pyramid Position Encoding Generator (PPEG) combined with an alternative Transformer architecture which approximates the self-attention mechanism the Nyströmformer (Xiong et al., 2021). GTP posits that combining ViTs along with graphs can lead to a more efficient approach. They build a graph from the extracted feature vectors and then passes the graph through a GCN and pooling layer before passing this to a transformer layer with the associated positional encodings.

Transformer methods have been shown in ablation studies to have a positive impact on model performance for a variety of tasks including glioma subtyping (M. Lu et al., 2021), metastasis detection (Shao et al., 2021), lung cancer subtyping (Shao et al., 2021; Zheng et al., 2021) and kidney cancer subtyping (Shao et al., 2021). Although some works (Shao et al., 2021) alter the structure of the positional encoding the majority of Transformer based approaches assume that the existing sinusoidal encoding approach proposed in (Dosovitskiy et al., 2021) is sufficient to include the required global context. While the Transformer proposals show they can outperform other existing approaches such as CLAM (M. Y. Lu et al., 2021) unfortunately there is not currently a comparison between the different transformer methods on the same tasks(s) .

### 3.1.2. Context-aware methods

To remedy the lack of context present in the bag of instance approaches described above several different methods outside of the Transformers have attempted to consider the spatial arrangement of instances as part of their pipeline.

Neural Image Compression (NIC) (Tellez et al., 2019) uses a representation learning approach. By training a tile encoder with a GAN and re-arranging the extracted feature vectors with the same spatial arrangement as the original tiles in the WSI they are able to compress the original WSI into a format which a CNN can hold in memory. Once extracted for each WSI this compressed image representation is then used to train a standard CNN architecture. This process assumes that by passing the instances through the GAN, the spatial relationships between them are preserved in the deep feature space.

Context-Aware CNN (Shaban et al., 2020) uses a similar idea but instead of just spatially re-arranging the instances in the same way they also use an attention block with the deep feature cube to include an encoding of the spatial context. This feature cube, with the spatial context encoded, is then passed to a classification CNN as in NIC. While Context-Aware CNN was only tested on HPFs of order $10^3 \times 10^3$ pixels it is very similar to NIC in terms of the overall pipeline and there's nothing preventing the approach from being applied at the WSI level.

Another approach to the same idea is Streaming CNN (Pinckaers et al., 2020). Here the authors attempt to train a CNN with arbitrary input image size by streaming the input image through the model in a series of large tiles and using gradient checkpointing to reduce the memory required to store the activations. By aggregating the gradients over large tile sizes e.g., 4096 or 8192 pixels they are able to train an end-to-end model for WSI classification for regressing to the PAM50 score (a measure for tumour growth) and for classifying metastases in breast cancer without aggregating a bag of instances.

The final paradigm is to not just incorporate an awareness of the spatial arrangement of instances but also of their hierarchical relationship as well into the overall pipeline. WSIs and other very high-resolution image formats use an image pyramid structure to store the image at different resolutions so they can display the region required at different magnifications. Similar to multi-resolution approaches one proposal (Jewsbury et al., 2021) splits HPF regions with a quadtree approach to create a bag of instances at different magnifications.

CellMaps (AlGhamdi et al., 2021) is another method for representation of histology images, which uses the cellular density of given image to represent the entire WSI. It can represent various cell types, each of which is corresponded in an image layer. The size of the CellMaps can be compressed to the desired level that the model/algorithm can handle, while all the relevant information is kept intact while reducing the image size. Besides, this representation captures the spatial information of cellular level details from the original image. AlGhamdi *et al*. (AlGhamdi et al., 2021) show that the prediction performance is improved when model is trained with the CellMaps representation, comparing with the raw H&E images.

### 3.2. Graph aggregation

In most of the aggregation approaches we have looked at in other sections, the instances which we are aiming to aggregate are treated as a collection of entities that have no specific relationship to each other beyond belonging to the same slide. The instance positions in the image could be permuted arbitrarily without affecting the aggregated prediction for the sample. However, these instances are patches, cells or other histological entities which have a spatial relationship to each other defined by their relative positions in a physical sample. Pathologists often place a lot of importance to the context in which biological features appear when trying to understand and diagnose a sample. The appearance of nearby tissue regions, or how a region of tissue fits into a larger biological structure, are important but are often not well accounted for in aggregation approaches, meaning useful information from the spatial relationships of the instances is lost.

Graph neural network approaches preserve these spatial relationships, by modelling the tissue as a graph of instances (in Graph terminology, the individual instances are referred to as nodes). This allows context from local neighbourhoods to be used when learning instance scores or representations, and can be used to inform global aggregation by identifying important nodes based on graph structure (W. Lu et al., 2022).

There are several excellent review articles covering graph neural networks in general (Wu et al., 2021) and GNNs in computational pathology (Ahmedt-Aristizabal et al., 2022), which we refer the reader to for an in-depth survey of GNN techniques. Here, we will present an overview of graph neural networks from the viewpoint of aggregation, focussing on the aspects most relevant in that context.

### 3.2.1. Graph representation

Let $G = (V, E)$ denote a graph, where $V$ and $E$ are the sets of nodes and edges respectively. Each node $v \in V$ is associated with a feature vector $F_v$. In the context of CPath, each node $v$ is often a histological entity such as a cell (J. Wang et al., 2019). It may also be a representation of a tissue region, such as a patch, or a cluster of patches or cells [4,45]. Some methods have also used pixel-based clustering methods such as SLIC to generate the nodes (Pati et al., 2020).The features $F_v$ will describe characteristics of the cell or tissue region. In general, graphs often also have associated edge features $F_e$, though this is less common in the context of CPath.

In learning instance (node) representations, GNNs aggregate information from a local neighbourhood, as illustrated in Figure 4a. The way in which this is done differs depending on the type of GNN it is.

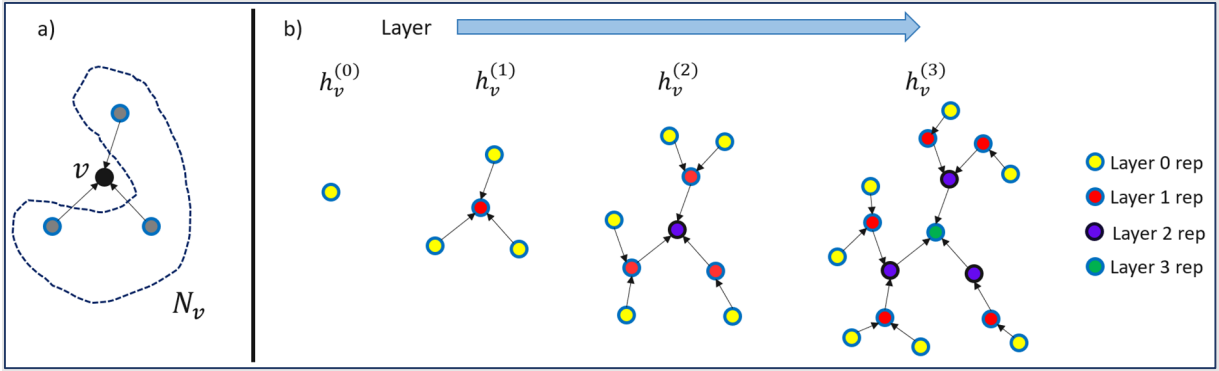

*Figure 4. a) Local neighbourhood of a node. b) Aggregation of information over a progressively larger region in successive layers of a spatial GNN*

A node $v$ in a graph has a local neighbourhood $N_v$ which is the set of all nodes to which it is connected by edges. Aggregation at a representational level usually occurs through messages passed to a node from the nodes in its neighbourhood.

The most general form of a spatial convolutional GNN is:

$$h_v^{(k)} = f_{\theta_k}\big(h_v^{(k-1)}, h_u^{(k-1)} | u \in N_v\big) \qquad (5)$$

A typical CGNN will have a small number (usually <10) of such layers, where at each layer the representation at a node aggregates information from a steadily larger region of the graph, as illustrated in Fig 3b. Depending on the connectivity of the graph and size of the individual instances, this can end up aggregating information from quite a large region of a WSI.

There has been a proliferation of suggestions in the literature for the form of equation (5), each trying to incorporate some specific intuition or satisfy some mathematical requirement on how a graph convolutional layer should be.

The EdgeConv graph convolution was introduced in (Y. Wang et al., 2019), and focusses on differences between the central node and its neighbours. It has been used for example in HER2 status prediction in [45].

In (K. Xu et al., 2019) the authors show that there are some graph structures that popular GNN variants cannot distinguish between, and propose the Graph Isomorphism network (GIN) to address this. This form allows for a weighting between the importance we place on the information in the central node, compared to the information aggregated from surrounding nodes in its neighbourhood $N$. It has been applied to breast cancer subtyping in (Pati et al., 2020).

Furthering the theme of investigating the expressiveness and representational learning power of graph neural networks, in (Corso et al., 2020) the authors identify that commonly used functions to aggregate the messages from a node's neighbourhood fail to distinguish between different message sets. Their proposal to solve this, Principal Neighbourhood Aggregation (PNA) involves using multiple different aggregators, as well as some degree-dependent scaling, to define a message aggregation function which can distinguish between a far larger variety of message sets.

Another popular method is GraphSage [51], a distinguishing feature of which is a sampling of a fixed size set of neighbours during aggregation. This can be especially useful to make learning a highly connected network computationally tractable. GraphSage has been used for cancer grading in (J. Wang et al., 2019; Y. Zhou et al., 2019).

### 3.2.2. Global graph aggregation

The second level at which a GNN aggregates information occurs at a global level, in a way much more directly analogous to the WSI aggregation covered in other sections. Given a GNN that outputs node-level scores, a 'graph readout' function is used to provide a graph level prediction. As the goal is simply to calculate a single score from a set of instance scores/representations, here we are back in the context of many of the other aggregation methods covered in this survey. We can take the mean, max, use attention, and so on.

A key concept in aggregation is the notion of node importance. The choice of aggregation method reflects our assumptions on which nodes are important to the aggregated prediction. The simplest (and surprisingly often used) aggregation method is to take the mean of the instance scores. In this case, we are implicitly assuming all instances are equally important. Many global aggregation strategies incorporate an estimate of node importance in the form of weighting or selection on the instances, either based on the instance score (e.g max/top N aggregation), or based on some representation of the instance (e.g attention based methods).

The spatial structure and relationships built into a graph representation can help us with this, by allowing us to define topological measures of importance on the graph structure. There has been much work in the literature on metrics to quantify the structural importance of a node in a graph, and for an in-depth review we refer the reader to (Lalou et al., 2018; Landherr et al., 2010). The most common approaches to global aggregation in CPath thus far have been simple aggregators such as the mean, as used in [45], [4], or attention-based approaches such as that in [48]. Another example of a learnable global aggregation is in [49], where node level representations output by a GNN are simply concatenated, and an MLP with learnable weights produces the global graph score.

### 5.2.3. Graphs In Computational Pathology

Here we will look in brief at a few examples of graph based approaches in CPath which illustrate the information aggregation aspects of graph-based methods particularly well. For a more general review of graph-based methods in CPath, we refer readers to [47].

Graph approaches in CPath usually consist of the following steps. 1. Entity detection/definition, 2. Feature embedding, 3. Graph construction, 4. GNN model training and prediction, 5. Model interpretation (for example using GNNExplainer (Ying et al., 2019)).

A good example of the different levels at which a graph based approach can aggregate information can be found in the SlideGraph+ [45] approach to building graphs for HER2 status prediction on WSIs. In this approach, the basic entities are image patches, which are represented by resnet50 (imagenet pretrained) features. Patches are clustered based on proximity in both position and feature space to form the graph nodes, which have the mean position/feature representation of the patches in the cluster. A further stage of aggregation then occurs during the learning of the graph representation of the node, and depending on if problem is a node or whole graph prediction task this could undergo a final stage of global aggregation such as a mean of node scores. The information in the slide is aggregated in stages: patches -> patch clusters -> graph aggregation over neighbouring patch clusters (nodes) -> global aggregation of node scores into a slide score.

A similar approach is Hact-Net, applied to both breast cancer subtyping (Pati et al., 2020), and gleason grading (Anklin et al., 2021). In this method, tissue regions are generated using a SLIC super-pixel approach. Texture features in these regions are combined with features from a cell-graph embedding of the cells in each region. Then a larger tissue graph is constructed using the tissue regions as the nodes.

Again, this is an excellent example of aggregation at multiple levels: Cells and the surrounding tissue pixels are aggregated into superpixels, in which one level of graph aggregation takes place. The superpixels are in turn used to build a graph, in which aggregation at a higher level occurs.

In other examples of graph-based approaches in CPath, in (J. Wang et al., 2019) cell graphs using morphological and local textural features are used for scoring of prostate cancer Tissue Micro-array (TMA) cores. Another grading application on, colorectal cancer, can be found in (Y. Zhou et al., 2019). A cell graph built on larger images is achieved by subsampling the detected cells in an image in a way that produces a representative sampling of nuclei across the image, from which a cell graph is then built.

## 4. Evaluation and comparative analysis

There are some challenges in CPath that pose the generalizability and verifiability crises. These challenges make it difficult for a single aggregation method to outperform the other methods in all CPath problems. Below we describe the two problems:

1. Generalizability: In routine clinical practice, tissue block is sectioned at multiple levels which are then mounted onto multiple slides. However, researchers in CPath are usually provided with one or two WSIs for each case. A major challenge is whether the data being provided (i.e., 1 or 2 images per case) contains the information needed for the downstream analysis. It is likely that the amount of information varies when providing different cohort for the same problem, due to the fact that the provided slide (or two) for each case does not necessarily always comprise the same information. The likelihood of such data variability is even higher when the tumour size is large, as the number of slides that can be sectioned is larger.

2. Verifiability: One of the objectives of the utilisation of ML techniques is to find novel biological insights, which has led some researchers to provide heatmaps showing some tissue regions playing significant roles in the downstream analysis. However, verifying whether the network/algorithm always picks the regions that are actually the causation of the problem being studied remains a challenge. It is possible that the "hot" tissue regions are just meaningless signals that correlated with the trained ground truth. One may argue that it is easy to identify whether the "hot" regions picked by the network are meaningful. This is true with problems like tumour region classification. In contrast, it is challenging when the aim of the study is to predict survival or molecular subtypes. In the absence of localized ground truth (i.e., particularly which tissue regions are directly the actual signals to some CPath problems), verifiability remains a challenge in this domain.

These challenges make it difficult to suggest a standardized solution for the CPath problems. Each problem with its own data can be analyzed with a specific computational solution, including the aggregation method. In the literature, there are a few studies that presented a comparison between different aggregation methods. For example, Laleh *et al*. (Laleh et al., 2022) compare the performance of several aggregation methods, including MIL-based methods and simple weakly-supervised (WS) methods. These methods are tested/evaluated against six different CPath problems. Their results show that the simple patch-based methods outperform the MIL-based methods.

In contrast, Zeng *et al*. (Zeng et al., 2022) present an opposite finding. The simple patch-based methods show the worse performance, comparing with the MIL-based methods. Such a contradiction proves the generalizability crisis in the CPath, where it is challenging for a single method to generally outperform in all problems and experimental settings. It is likely that for each CPath problem, a different aggregation method is more suited, based on its assumption and experimental setups (e.g., amount of available data). The top method for that problem might fail when the data or the problem is a different one.

Primarily, the experimental analysis evaluates the overall performance of the prediction accuracy, the generalization, and verification of the prediction for the downstream prediction tasks, as performed by Laleh *et al*. (Laleh et al., 2022). We identify aggregation as an essential part of CPath applications. The goal of the aggregation method is to combine all the processed information available in a WSI into a final score or category. We argue that factors like aggregation, the type, nature, and amount of input data, features and the underlying machine learning approach impact the overall performance of slide-level predictions. Laleh *et al*. (Laleh et al., 2022) partially considered this part of the comparative evaluation in their benchmarking study. However, they did not make a fair comparative analysis of different aggregation methods in the MIL-based setting. Therefore, we conduct a case study of head-to-head comparative analysis of various benchmark aggregation methods.

### 4.1. Case study: Fair comparative analysis of popular aggregation methods

It is crucial to conduct a fair head-to-head comparisons to evaluate the performance of different aggregation methods by fixing all the components within the pipeline apart from the aggregation method they choose. Laleh *et al*. (Laleh et al., 2022) conducted comprehensive experiments for benchmarking different end-to-end CPath pipeline. However, their results might not be appropriate for comparing of different aggregation methods. In their paper, ResNet (He et al., 2016), EfficientNet (Tan & Le, 2021) and ViT (Dosovitskiy et al., 2021) were all pre-trained on ImageNet and then fine-tuned on the target datasets, while MIL (Campanella et al., 2019; Dietterich et al., 1997), AttMIL (Ilse et al., 2018) and CLAM (M. Y. Lu et al., 2021) used a ResNet feature extractor, which was only pre-trained on ImageNet. Secondly, different methods in their work used different backbone networks. The ResNet approach in their paper used ResNet-18 structure, EfficientNet used efficientnet-b7 structure, ViT used vision transformer and the MIL, AttMIL and CLAM used ResNet-50 structure. Moreover, based on the published code for CLAM, feature extraction uses the features extracted after the third ResNet block, whereas the standard procedure is to use the features generated before the final classification layer. There are so many variables in existing approaches that had not been controlled and may introduce bias into the comparisons. Therefore, we conduct a case study where we attempt to fairly compare different aggregation methods. Moreover, our case study shows that there are many components within an end-to-end framework which can impact the overall performance of the algorithm, sometimes even more than the aggregation method does.

### 4.1.1. HPV infection prediction in head and neck cancers

We chose the problem of Human papillomavirus (HPV) infection status prediction in head and neck cancers as a case study for comparing some popular aggregation methods. HPV infection status is an important biomarker in head and neck cancers which can affect the prognosis, survival and the treatment selection. While immunohistochemistry and PCR are the gold standard for HPV infection diagnosis in the clinical practice, there have recently been a few attempts in the CPath community to solve this problem from analysing digital H&E slides (Kather, Schulte, et al., 2019; Klein et al., 2021).

We believe this problem is ideal as a case study for comparing aggregation methods for several reasons. First of all, like other CPath problems, it requires aggregating predictions from tiles extracted from WSIs. Secondly, clinicians have categorised many histological differences between HPV+ and HPV- H&E slides (Westra, 2012). However, none of these is distinct enough to become a gold standard for human pathologist to reach a diagnosis. It is, therefore, quite possible that the data we provide to the algorithm contains information to enable discrimination between two types of carcinomas, as well as posing some challenges.

### 4.1.2. Experimental Settings

We chose 6 different aggregation methods for our case study, which are majority voting, mean pooling, max pooling, median pooling, attention module (Ilse et al., 2018) and CLAM (M. Y. Lu et al., 2021). For majority voting, we use weak labelling to train a classification model, and use the proportion of positive tiles predicted as the prediction score. For mean, max and median pooling and attention module, each of these pooling methods was tested for generating either a WSI-level score or feature from the tile-level features, and then a model was trained using ranking loss (R. Wang et al., n.d.). For CLAM, we used the training and aggregation module of CLAM, while using different feature extractors which were in accordance with the comparable experiments.

The dataset we used was retrieved from the Head and Neck Squamous Cell Carcinoma cohort of The Cancer Genome Atlas project. The study by Campbell *et al*. (Campbell et al., 2018) provides us with the HPV infection status for these patients. We followed the same approach in Kather *et al*. (Kather, Schulte, et al., 2019) to exclude the data which were not qualified to include in our experiment. The 512x512 size patches were extracted under the 20X magnification within the tumour regions and then were resized to 224x224. Three-fold strong cross validation experiments were conducted where the dataset was split into 3 folds and each fold was used as training, validation and testing set in each of the 3 cross-validation experiments. The mean and standard deviation of the AUROC were reported over the 3-fold cross-validation experiments as the metric for evaluation.

### 4.1.3. Benchmark performance metrics

The area under the receiver operating characteristic (AUROC) was chosen to be the quantitative metric for evaluating the performance of different aggregation methods. The receiver operating characteristic (ROC) curve is widely used in machine learning research to evaluate the performance of classifiers. The ROC curve is plotted with the true-positive rates and the false-positive rates generated by setting different cut-off values to the predictions. The true-positive rate can be interpreted as the sensitivity, and (1- false-positive rate) can be interpreted as the specificity. The AUROC is the area under the plotted ROC curve, which is a quantitative metric for describing ROC curves and the performance of the algorithms. Better the algorithm is, closer the ROC curve would be to the upper-left point of the plot, and larger the AUROC would be. The ROC curve is also widely adopted in CPath community to analyse the clinical sensitivity and specificity of the proposed algorithms. Our case study also chooses AUROC as the evaluation metric.

### 4.1.4. Results

Table 2 lists 6 different aggregation methods and their corresponding results in terms of AUROC values, and the mean pooling and the attention module reported the best mean AUROC, while the mean pooling has a lower standard deviation compared to the attention module. All the experiments in Table 2 used ResNet-18 as the feature extractor, which was pretrained on ImageNet and then fine-tuned on the target dataset.

Table 3 shows two experiments which use the same aggregation method but different backbone networks, ResNet-18 and VGG-19. ResNet-18 backbone performed 0.06 better in AUROC than VGG-19 backbone. The results here show that different backbone networks used for feature extraction can also impact the performance of the whole pipeline.

Table 2. Results comparison between different aggregation approaches.

| Aggregation Approach | AUROC |
|---|---|
| Majority Voting | 0.8611±0.0061 |
| Mean Pooling | **0.8673±0.0062** |
| Max Pooling | 0.8092±0.0064 |
| Median Pooling | 0.8515±0.0186 |
| Attention (Ilse et. al. (Ilse et al., 2018)) | 0.8673±0.0148 |
| CLAM (Lu et. al.(M. Y. Lu et al., 2021)) | 0.8117±0.0239 |

Table 3. Results comparison between different backbone networks.

| Aggregation Approach | Backbone | AUROC |
|---|---|---|
| Majority Voting | ResNet-18 | 0.8611±0.0061 |
|  | VGG-19 | 0.8040±0.0151 |
| CLAM | ResNet-18 | 0.8117±0.0239 |
|  | ResNet-50 | 0.8385±0.0104 |

Table 4 presents the results achieved by using the same aggregation method, the same backbone network structure, but with different ways of pretraining. The ImageNet pretrained ResNet-18 only uses the ImageNet pretrained weight for feature extraction, while the fine-tuned ResNet-18 uses ImageNet pretrained weight and then fine-tuned on the target dataset. We can see from this table that the difference caused by between different aggregation methods can go up to 0.06 in AUROC, while different pretraining approach can impact the performance up to 0.14 in AUROC. This experiment shows that some components within the CPath pipeline other than the aggregation method can actually impact more on the final performance. Therefore, we believe the researchers should be extremely careful when reaching a conclusion that one aggregation method is better than another.

Table 4. Results comparison between differently trained backbone networks.

| Aggregation Approach | AUROC (Fine-tuned ResNet-18) | AUROC (ImageNet Pretrained ResNet-18) | Difference |
|---|---|---|---|
| Majority Voting | 0.8611±0.0061 | N/A | N/A |
| Mean Pooling | 0.8673±0.0062 | 0.731±0.0589 | 0.14 |
| Max Pooling | 0.8092±0.0064 | 0.7419±0.0287 | 0.07 |
| Median Pooling | 0.8515±0.0186 | 0.7683±0.0108 | 0.08 |
| Attention (Ilse et. al. (Ilse et al., 2018)) | 0.8673±0.0148 | 0.7227±0.0638 | 0.10 |
| CLAM (Lu et. al.(M. Y. Lu et al., 2021)) | 0.8117±0.0239 | 0.7035±0.0356 | 0.14 |
| **Maximum Difference** | 0.06 | 0.06 | 0.14 |

## 5. Discussion, conclusions and future directions

We analysed a variety of computational pipelines for predictive modelling in CPath and grouped them in terms of data and computational frameworks. CPath offers data with different levels of details and contextual information, from pixel to patch and cell or gland to tissue phenotypes. Consequently, the computational frameworks originated in relation to the scheme of representing different levels and types of data and the contextual information, e.g., from individual patches or cells to their connectivity through graphs. Besides, the output aggregation methodologies came in as simple pooling, data-driven like machine learning, and clinical rules. In addition, the computational resources, time, and related costs have their impacts on the modelling.

Each of these components of computation pipelines contributes to the success within a given problem frame, but none of it comes without challenges or trade-offs and can be called a single best solution for all CPath problems. It's critical to identify the input and the main goal, design the best solution for the given problem, and define metrics for the success. The significant most metrics are the predictive performance and interpretability or explainability of the predictions. In terms of predictive performance, e.g. in the diagnostic applications the ultimate goal may be a performance comparable to existing systems i.e., clinical practice, and even further improvement through an objective analysis. Both these, the predictability and explainability, are open research problems so far and offer multidisciplinary contributions to impactful solutions and novel insights.

The case study conducted in this paper validates the above argument. Our case study explicitly shows that there are at least two aspects in a CPath pipeline other than the aggregation method. These are different backbone networks and pretraining approaches for feature extractor, which can have an impact on the overall performance of the aggregation workflow. In a fair comparative analysis of aggregation method, it is essential to control all the variables of experiments other than the aggregation method chosen. For those who want to choose the best aggregation method for their research, it is important to consider the pipeline, the clinical explainability and the problem in question. This is because the performance of different aggregation methods varies depending on many aspects, making it challenging to determine the most optimal one. More importantly, we believe all the CPath researchers should also bear in mind that the quantitative metrics is not the only thing to pursue. Instead, the generalizability, verifiability of the CPath approach, the biological interpretability and explainability, the ability to generate localised predictions and novel insights into the aetiology of diseases might be more important than a high accuracy. Asif et al. (Asif et al., 2021), in their review article, urged the need of rigorous testing of AI model as one of CPath challenges and limitations associated with the AI development lifecycle.

Our case study shows (both in Table x1 and x2) that fine-tuning has a higher impact on performance than the fixed ResNet features. It allowed both pooling and data-driven aggregation with attention MIL giving the same accuracy. It also suggests that attention MIL does not have an added value in predicting the status of the HPV virus as compared to the mean pooling. CLAM (M. Y. Lu et al., 2021) relies only on transfer learning and data-driven aggregation. The authors choose to build a data- and resource-efficient pipeline by excluding fine-tuning and using less than 100% data for training efficiency in some experiments. Though, the feature extraction also has additional costs in extracting and storing feature vectors of all the patches. Their aggregation method performs better with ResNet50 features as they aimed. CALM workflow with ResNet50 feature vectors works better than fine-tuned ResNet18 features. However, it requires further improvement to make the fine-tuning non-essential for aggregation in CPath as compared to RankMIL workflow(R. Wang et al., n.d.) as shown in our case study and a naïve MIL workflow for six different benchmark problems in (Laleh et al., 2022).

Several bottom-up workflows for WSI level prediction have been developed recently. A major advantage of bottom-up approach over the top-down approach is its better explainability and interpretability. Predictive modelling of such methods require detailed cell and region-level annotations but often work well with small amount of data as compared to top-down modelling. Diao et al. in (Diao et al., 2021) have demonstrated use of human interpretable features to predict diverse molecular signatures (AUROC 0.601–0.864), including expression of four immune checkpoint proteins and homologous recombination deficiency, with performance comparable to top-down approach. Ho et al. in (Ho et al., 2022) got similar findings with their gland segmentation based bottom-up approach for the screening of colorectal cancer. Yamashita et al. in (Yamashita et al., 2021) has demonstrated better prediction of MSI status than naïve MIL top-down approach (Kather, Pearson, et al., 2019). For slide-level MSI prediction, Park et al. in (Park et al., 2022) have compared a top-down approach with a bottom-up approach in which they have combined features from multiple objects and levels of inputs including tissue phenotypic, cells, and glands. The top-down approach produced better AUROC scores whereas the bottom-up produced explainable features to verify differentiating features with expert knowledge.

AI algorithms are prone to biases particularly introducing positive bias when developed and validated in siloed manners that results in deteriorated performances on external cohorts revealing generalisation deficiencies. This commonly occurs as developers have control on establishing validation cohorts and readout experiments. Therefore, it is crucial to evaluate the generalisation of AI algorithm independently across different patient populations, pathology labs, digital pathology scanners, reference standards derived from global panel (Bulten et al., 2022). Data bias, quality and reproducibility of the results are also key challenges in the AI development life cycle (Asif et al., 2021).

Global AI competitions have been an effective approach to overcome the pitfalls of soiled development by crowd sourcing the development of the performant algorithms. These competitions can also overcome generalisation issues if they implement an independent evaluation appropriately, such as in a recent Prostate cANcer graDe Assessment (PANDA) competition (Bulten et al., 2022), which is a single largest competition in pathology to date. They were able to fully-reproduce top-performing 15 algorithms and externally validated their generalisation to independent US and EU cohorts and compared them with the reviews pf pathologists.

In PANDA challenge, a total of 1,010 teams, consisting of 1,290 developers from 65 countries participated and submitted at-least one algorithm of total 34,262 versions. The winner and the most leading teams, adopted an aggregation approach in which a sample of smaller tiles are processed by CNNs and predictions are concatenated in the final classification at WSI-level, without requiring any detailed region/pixel-level annotations. Two of the exciting findings include the similar to and higher statistically significant agreement of algorithm with the uropathologists and higher sensitivities for tumour identification than representative pathologists on external validation subsets of both EU and US, respectively.

All three workflows alongside different aggregation approaches have advantages and challenges associated with them. The bottom-up approaches proposed in (Diao et al., 2021; Park et al., 2022) combined tissue phenotype based workflow (third workflow in Figure 2) with objects (cells and glans) level workflows, which is likely to be explored further in subsequent studies as its potential and comparative advantages are unconclusive. The emerging trends in CPath combine MIL, attention/transformer mechanisms, graph representation, and learning for better accuracy, generalization, and interpretation of data for various clinical applications (H. Chen et al., 2022; R. J. Chen et al., 2022; Guan et al., 2022; Javed et al., 2022; Kosaraju et al., 2022; Zheng et al., 2022). Next-generation CPath workflows could have priority research problems concerning transparency and interpretability of predictions in MIL/top-down workflows, versus clinical-grade/better predictability in interpretable bottom-up workflows.

Other factors may include data efficiency, which assesses the amount of data and the labelled data needed for robust and efficient machine learning. The efficiency of computational resources, the hardware, and the turnaround time required to reach the final WSI-level prediction becomes important when implemented in a real-world setting. To accelerate predictive modelling for CPath solutions to the next level may also require considering the notion of learning paradigm, e.g., end-to-end learning and self/unsupervised learning and new ways of modelling attention mechanism and data loading pipeline for selection of most relevant image tiles for training. It is, however, expected that future research and development will produce more data and allow a broader evaluation to rank approaches for different CPath use cases and performance criteria.

**Acknowledgement**

Authors are grateful to members of the Tissue Image Analytics (TIA) centre for their feedback on some of the initial ideas for this paper.

**Authors Contributions**

Mohsin Bilal: Conceptualization, Writing- Original Draft, Investigation, Writing- Reviewing and Editing, Visualization, Project administration.: Robert Jewsbury: Writing- Original Draft, Investigation, Visualization.: Ruoyu Wang: Methodology, Software, Validation, Writing- Original Draft.: Hammam Alghamdi: Writing- Original Draft, Investigation, Visualization: Amina Asif: Investigation, Reviewing and Editing.: Mark Eastwood: Writing- Original Draft, Investigation, Writing- Reviewing and Editing, Visualization.: Nasir Rajpoot: Supervision, Conceptualization, Investigation, Reviewing and Editing, Funding acquisition.